\documentclass[runningheads]{llncs}
\usepackage{url}
\usepackage{blindtext,alltt}
\usepackage[hang,small,bf]{caption}
\usepackage[utf8]{inputenc}
\usepackage{xspace}
\usepackage{amsmath}
\usepackage{amsmath}
\usepackage{amssymb}
\usepackage{bm}
\usepackage{url}
\usepackage{subfigure}
\usepackage{threeparttable}
\usepackage{bbm}
\usepackage{enumitem}
\usepackage{marvosym}
\usepackage{bbm}
\usepackage{ifsym}
\usepackage{graphicx}
\usepackage{amsmath}
\usepackage{amssymb}
\usepackage{enumitem,amssymb}
\usepackage{algorithm}
\usepackage{booktabs} 
\usepackage{enumerate}
\usepackage{algorithmic}
\usepackage{subfigure}
\usepackage{caption}
\usepackage{threeparttable}
\usepackage{booktabs}
\usepackage{float}
\usepackage{threeparttable}
\usepackage{multicol}
\usepackage{multirow}
\usepackage{subfigure} 
\usepackage[colorlinks,
            linkcolor=blue,
            anchorcolor=blue,
            citecolor=blue]{hyperref}
\usepackage{geometry}
\usepackage[skip=9pt,font=scriptsize]{caption}

\usepackage{blindtext}
\usepackage[T1]{fontenc}
\usepackage[utf8]{inputenc}
\usepackage{cite}
\begin{document}
\pagestyle{empty}  
\thispagestyle{empty} 
\title{\Large\bfseries  Dopamine Transporter SPECT Image Classification for Neurodegenerative Parkinsonism via Diffusion Maps and Machine Learning Classifiers}

%
%
\author{Jun-En Ding\inst{1}\textsuperscript{(\Letter)} \and
Chi-Hsiang Chu\inst{2} \and
Mong-Na Lo Huang\inst{3} \and \\ Chien-Ching Hsu\inst{4}\textsuperscript{(\Letter)}}
\authorrunning{D. Jun-En et al.}
%
\institute{Research Center for Information Technology Innovation, \\
    Academia Sinica, Taipei, Taiwan
    \email{ding1119@citi.sinica.edu.tw}\and
Department of Statistics, National Cheng-Kung University, Tainan, Taiwan 
    \and
    Department of Applied Mathematics
    National Sun Yat-sen University, \\
    Kaohsiung, Taiwan \and
    Department of Nuclear Medicine, Kaohsiung Chang Gung Memorial Hospital,\\
Chang Gung University College of Medicine, Taiwan\\
}

\maketitle              
\begin{abstract}
Neurodegenerative parkinsonism can be assessed by dopamine transporter single photon emission computed tomography (DaT-SPECT). Although generating images is time consuming, these images can show interobserver variability and they have been visually interpreted by nuclear medicine physicians to date. Accordingly, this study aims to provide an automatic and robust method based on Diffusion Maps and machine learning classifiers to classify the SPECT images into two types, namely Normal and Abnormal DaT-SPECT image groups. In comparison with deep learning methods, our contribution is to propose an explainable diagnosis process with high prediction accuracy. In the proposed method, the 3D images of $N$ patients are mapped to an $N \times N$ pairwise distance matrix and are visualized in Diffusion Maps coordinates. The images of the training set are embedded into a low-dimensional space by using diffusion maps. Moreover, we use Nystr$\ddot{o}$m's out-of-sample extension, which embeds new sample points as the testing set in the reduced space. Testing samples in the embedded space are then classified into two types through the ensemble classifier with Linear Discriminant Analysis (LDA) and voting procedure through twenty-five-fold cross-validation results. The feasibility of the method is demonstrated via Parkinsonism Progression Markers Initiative (PPMI) dataset of 1097 subjects and a clinical cohort from Kaohsiung Chang Gung Memorial Hospital (KCGMH-TW) of 630 patients. We compare performances using Diffusion Maps with those of three alternative manifold methods for dimension reduction, namely Locally Linear Embedding (LLE), Isomorphic Mapping Algorithm (Isomap), and Kernel Principal Component Analysis (Kernel PCA). We also compare results using 2D and 3D CNN methods. The diffusion maps method has an average accuracy of 98\% for the PPMI and 90\% for the KCGMH-TW dataset with twenty-five fold cross-validation results. It outperforms the other three methods concerning the overall accuracy and the robustness in the training and testing samples. 

\end{abstract}
\section{\large\bfseries Introduction}

Parkinson’s disease (PD) is a neurodegenerative disorder, and its pathological feature is the loss of dopaminergic neurons in the substantia nigra \cite{Kalia2015}. Currently, the diagnosis of PD is mainly based on the clinical symptoms (tremor at rest, rigidity, bradykinesia, gait disturbance) and the response to medication (levodopa or dopamine agonist). Parkinsonian syndromes refer to some diseases with clinical symptoms which is similar to PD. The etiologies of Parkinsonian syndromes may also be related to the degeneration of dopamine neurons (such as multiple system atrophy, progressive supranuclear palsy, dementia with Lewy bodies, and corticobasal degeneration) or non-neurodegenerative (such as essential tremor, secondary parkinsonism related to hydrocephalus, stroke, drugs, toxins, trauma, brain tumor, or infection)
\cite{Keener2016,Hayes2019}.

Dopamine transporters (DaT) are located on the presynaptic dopaminergic nerve terminal and play an important role in regulating extracellular dopamine level via reuptake dopamine into the nerve terminal. The loss of dopaminergic neurons in neurodegenerative parkinsonism leads to the reduction of DaT. Many radiotracers for single photon emission computed tomography (SPECT) such as [$^{99m}$Tc]TRODAT-1 \cite{Kung2007}, $^{123}$I-beta-CIT, $^{123}$I FP-CIT, which can bind specifically with DaT, were developed to diagnose PD and neurogenerative parkinsonism.

There are studies for diagnosis of Parkinson's disease based on features extracted from SPECT images, such as shape and volume, etc.,  combining with statistical tests or machine learning classifiers \cite{ref3_,SPECT5967908}.
However, features like the shape of the abnormal striatum may be quite different and in certain cases the features cannot be captured correctly, especially when most images require high-dimensional analysis, the proposed analysis can be time consuming and may have a large estimation error due to noisy image. 

In recent years, due to rapid advancements in data storage and hardware technology, neural network methods have become a powerful technique for prediction and image feature space extraction. One of the most well-known neural networks is the convolutional neural network (CNN). CNNs have been widely applied for medical image analysis, including MRI and fMRI, well-known CNN models using transfer learning architecture such as
AlexNet\cite{Krizhevsky12imagenetclassification}, VGG-16\cite{Simonyan2015}, VGG-19 and Deep Convolution Network (DCNN). The series of VGGNet was first proposed VGG-16 in 2014-ILSVRC competition followed by VGG-19 as two
successful architectures on ImageNet. VGG-16 and VGG-19 use different frameworks, and their models make an improvement on AlexNet by replacing large kernel-sized filters with multiple small kernel-sized filters resulting in 13 and 16 convolution layers for VGG-16 and VGG-19 respectively. In addition, more and more deep learning classifications of $^{123}$I SPECT scans use transfer learning from deep neural networks pretrained on nonmedical images \cite{transfer learning}. 

However, the deep learning results are often require a large training set to improve their classification accuracy. In general, biomedical images have a high dimnesion feature space, which further complicates the classification process. Consequently, it is desirable to perform dimension reduction to improve the efficiency of the training and testing process.  Many well-known non-linear dimension reduction methods such that Kernel PCA,  multidimensional scaling (MDS), Isometric feature mapping, and locally linear embedding were used for image classification and improve speed \cite{Faaeq2018}. 

In comparison with previous studies, we use a manifold learning methodology, namely the Diffusion Maps (DM) \cite{Coifman05geometricdiffusions} to perform dimension reduction of the SPECT image for classification purposes. Diffusion Maps do not require complicated parameter adjustment and therefore have a shorter training time and require fewer samples to capture more important information in low-dimensional space. In addition to classification or early diagnosis and visualization, a general standard specification for non-linear dimensionality reduction methods is spectral decomposition. Through mapping the images into a low-dimensional space with new coordinates corresponding to the eigenspace associated with the largest few eigenvalues, it is hoped that the points that have a relationship between samples can be as close as possible after dimensionality reduction, while still keep the initial data structure.
The method which refers to Diffuison Maps hopes to find the geometric description of the corresponding low-dimensional data through the diffusion process. In summary, we make three major contributions in this paper:

\begin{itemize}
  \item  
  We propose a robust method for images embedding associated with their striatum similarity and combine with the ensemble classifier.
  \item 
  We provide a diagnosis procedure for every patient with twenty-five voting to predict and construct an interpretable two-model confusion matrix.
  \item 
  We conduct extensive experiments on two real-world datasets. One is obtained from the benchmark of Parkinson's Progression Markers Initiative (PPMI) database \cite{Cummings2011,Quan2019}, and the other one is 
  Taiwan clinical database. In our method, we can use low-cost computing time than deep learning method and get better and more robust results.  
\end{itemize}

\section{Datasets of PPMI and Clinical Cohort}

\subsection{PPMI Dataset}

Data for this study were obtained from the PPMI database, a longitudinal, multicentre study to assess the progression of clinical features, imaging, and biologic markers on PD patients and healthy controls (HC). All the PD subjects were at an early stage of Hoehn and Yahr stage I or II at baseline. The diagnosis of PD was confirmed from the PPMI imaging core that the screening DaT-SPECT ($^{123}$I FP-CIT) is consistent with DaT deficit. Further details of the PPMI database can be found at (\href{http://www.ppmi-info.org}{\color{blue}{http:// www.ppmi-info.org}}). The PPMI dataset in our study contained 876 PD and 414 HC subjects.

We divide our KCGMH-TW samples into two classes (normal and abnormal). We perform analyses with data based on a single ($1\times128\times128\times3$, i.e., the center image in \textbf{Fig. \ref{fig:data_example}}) or three ($3\times128\times128\times3$, i.e., the three images of the middle row in \textbf{Fig. \ref{fig:data_example}}) from KCGMH-TW in our investigation. The PPMI ($1\times109\times91\times3$) dataset has been divided into two classes (HC and PD).

\subsection{Clinical Dataset of KCGMH-TW}

To attest the effectiveness of the proposed method in KCGMH-TW clinical dataset, we enrolled 730 patients who underwent [$^{99m}$Tc] TRODAT-1 brain SPECT between January 2017 and June 2019 in the Kaohsiung Chang Gung Memorial Hospital, Taiwan (KCGMH-TW). The Chang Gung Medical Foundation Institutional Review Board approved this retrospective study and waived the requirement for obtaining informed consent from the patients. Each patient was intravenously injected with a 925-MBq dose of [$^{99m}$Tc] TRODAT-1 (Institute of Nuclear Energy Research, Taiwan). [$^{99m}$Tc] TRODAT SPECT images were acquired using a hybrid SPECT/CT system (Symbia T; Siemens Medical Solution). SPECT images were obtained with 30s per step acquiring 120 projections over a circular 360 degree rotation using low-energy, high-resolution parallel-hole collimators. A $128\times 128$ matrix and a $\times1.45$ zoom were used. The CT images were acquired without contrast medium; they used the following parameters: 130 kV; 45mAs (Image Quality Reference mAs, CARE Dose 4D; Siemens Medical Solutions); rotation time, 1.5s; collimation, $2 \times 2.5$ mm. CT images were reconstructed to a $512 \times 512$ image matrix with a very smooth kernel, H08s (Siemens Medical Solutions) for SPECT attenuation correction. Raw SPECT data were reconstructed into transaxial slices using flash 3D (OSEM reconstruction method with 3D collimator beam modeling) with 8 subsets and 8 iterations and corrected with the H08s CT attenuation map. Images were smoothed using a 3D spatial Gaussian filter (fullwidth at half maximum, 6mm). The reconstructed transaxial slice thickness was 3.3 mm.

\begin{figure}[H]
    \centering
    \includegraphics[width=0.5\linewidth]{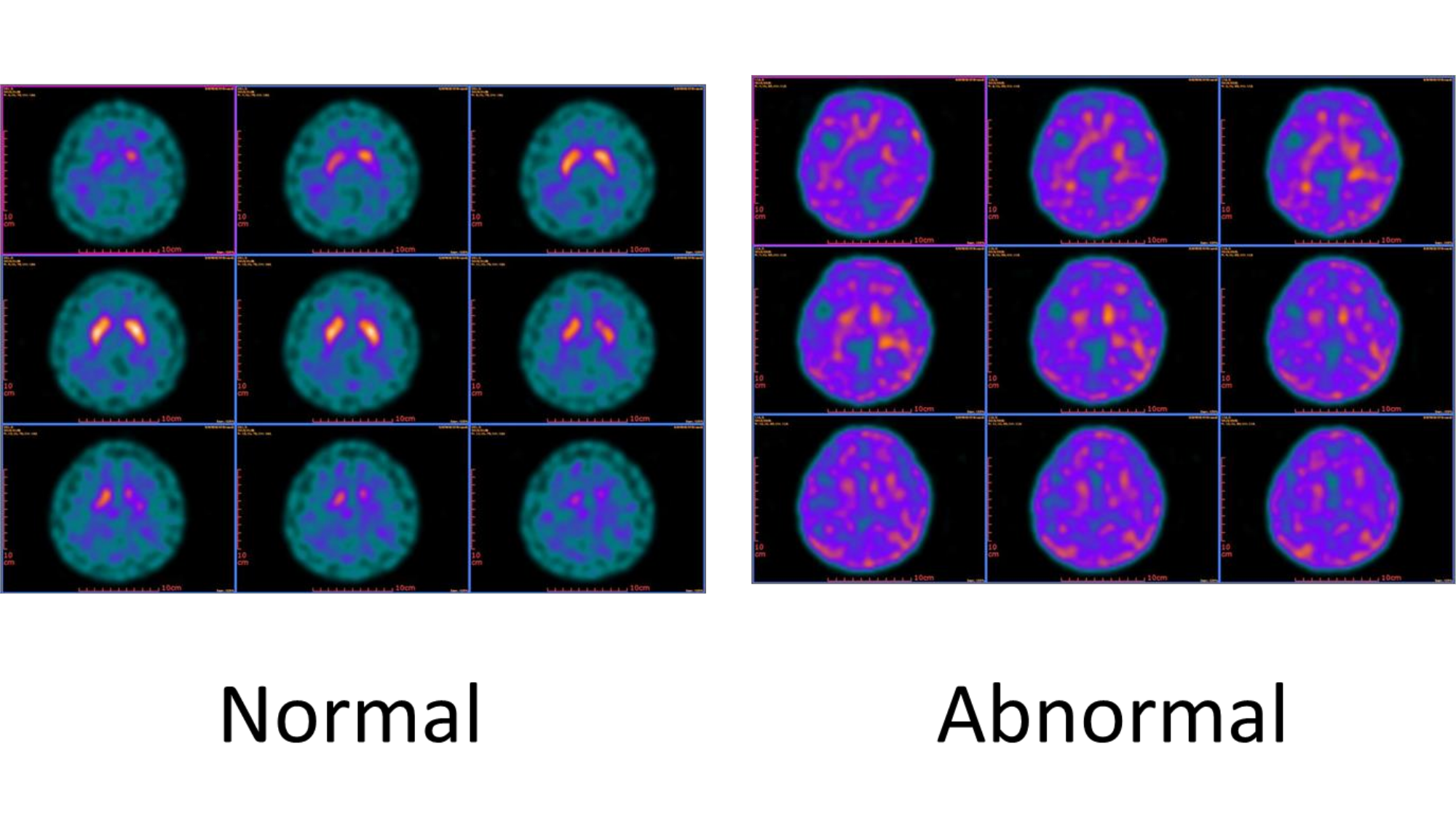}
    \captionsetup{font={normalsize}} 
    \caption{Nine consecutive transaxial images of dopamine transporter single photon emission computed tomography (DaT-SPECT) showing the whole striatal radioactivity were displayed in a $3 \times 3$ slices for visually interpreted. (\textbf{Left}) The normal images of DaT-SPECT show symmetrical, well-delineated comma-shaped radioactivity in the bilateral striata. (\textbf{Right}) The abnormal images of DaT-SPECT show reduced radioactivity of the bilateral striata (nearly equal to the background radioactivity of the brain). The putamen is usually more severely affected than caudate nucleus resulting in a circular or oval shape.}
    \label{fig:data_example}
\end{figure}

We select nine consecutive SPECT transaxial images showing the whole striatal radioactivity. Each patient has a separate 9 slices SPECT images, where it is more visible as shown in \textbf{Fig. \ref{fig:data_example}}. The pixel size of each cell is $128\times128\times3$ (RGB), so we take the middle best three slices of striatal images by nuclear medicine physicians \cite{SPECT5967908} and combine into a single image from the nine slices SPECT as our research target.

\subsection{Labeling Criterion}

All DaT-SPECT images were visually interpreted by three experienced board-certified nuclear medicine physicians according to Society of Nuclear Medicine practice guideline \cite{Djang2012}. The labeling criteria were established manually as follows. (1) Normal DaT-SPECT images: the normal striata on transaxial images should look crescent- or comma-shaped and should have symmetric well-delineated borders. 
(2) Abnormal DaT-SPECT images: the abnormal striata have reduced intensity on one or both sides, often shrinking to a circular or oval shape. The putamen is usually more severely affected than the caudate nucleus.

Blinded to patients' clinical information except age and gender, three nuclear medicine physicians visually interpreted the DaT-SPECT images independently. The final consensus result of normal or abnormal image was assigned if at least 2 physicians achieved an agreement. The PPMI dataset images ($1 \times 109 \times 91 \times 3 $) were divided into two classes (HC and PD). Similarly, the clinical  dataset images of KCGMH-TW were divided into two classes (normal and abnormal). The training/validation/testing dataset were summarized in \textbf{Table. \ref{tab:table}}. We analyzed KCGMH-TW dataset images based on single image ($1\times128\times128\times3$, i.e., the center image in \textbf{Fig. \ref{fig:data_example}}) or three ($3\times128\times128\times3$, i.e., the three images of the middle row in \textbf{Fig. \ref{fig:data_example}}).

\begin{table}[H]
\newcommand{\tabincell}[2]{\begin{tabular}{@{}#1@{}}#2\end{tabular}}
 \caption{Dataset statistics}
  \centering
  \begin{tabular}{lll}
    \multicolumn{3}{c}{}                   \\
    \toprule
    \cmidrule(r){1-2}
    \tabincell{c}{Dataset \\ (training/ validation /testing)}
        & Two classes     & Mean age \\
    \midrule
     \tabincell{c}{PPMI \\ (872/225/193)}  & \tabincell{c}{HC (n=414) \\ PD (n=876)}
      & -\\
    \cmidrule(r){2-3}  
    
    \tabincell{c}{KCGMH-TW \\(504/126/100)}     & \tabincell{c}{Normal (n=353) \\ Abnormal (n=377)} & \tabincell{c}{68.4  \\68.2}\\
    
    \bottomrule
  \end{tabular}
  \label{tab:table}
\end{table}

\section{Methodology}

In this study, we firstly propose the use of diffusion maps for dimension reduction of the data, and then find the corresponding classification methods for disease diagonsis. Accuracy of the proposed methodology is evaluated through a cross-validation procedure for the training samples and later for the testing samples. For making a cross-validation study, we split a total of 630 original data samples to normal and abnormal group. Then we randomly divide each group into five folds and cross combine them into 25 folds of paired data. And then each fold of partition to five folds according to  individual normal and abnormal group. The divided data will be training cross combination and validate until each combination sample has been trained. 
Our proposed method applies each training set and conducts Nystrom's out-of-sample expansion \cite{ScNormallar2012} new sample to project training space, which reduced space will be classified by optimal classifier in validation processing. Finally, we will add a new sample and fixed test set (n=100) for prediction, and each patient in this test set will have a total of twenty-five votes. 

\subsection{Training Sample Reduction via Diffusion Maps}

\begin{figure}[H]
    \centering
    \includegraphics[width=0.58\linewidth]{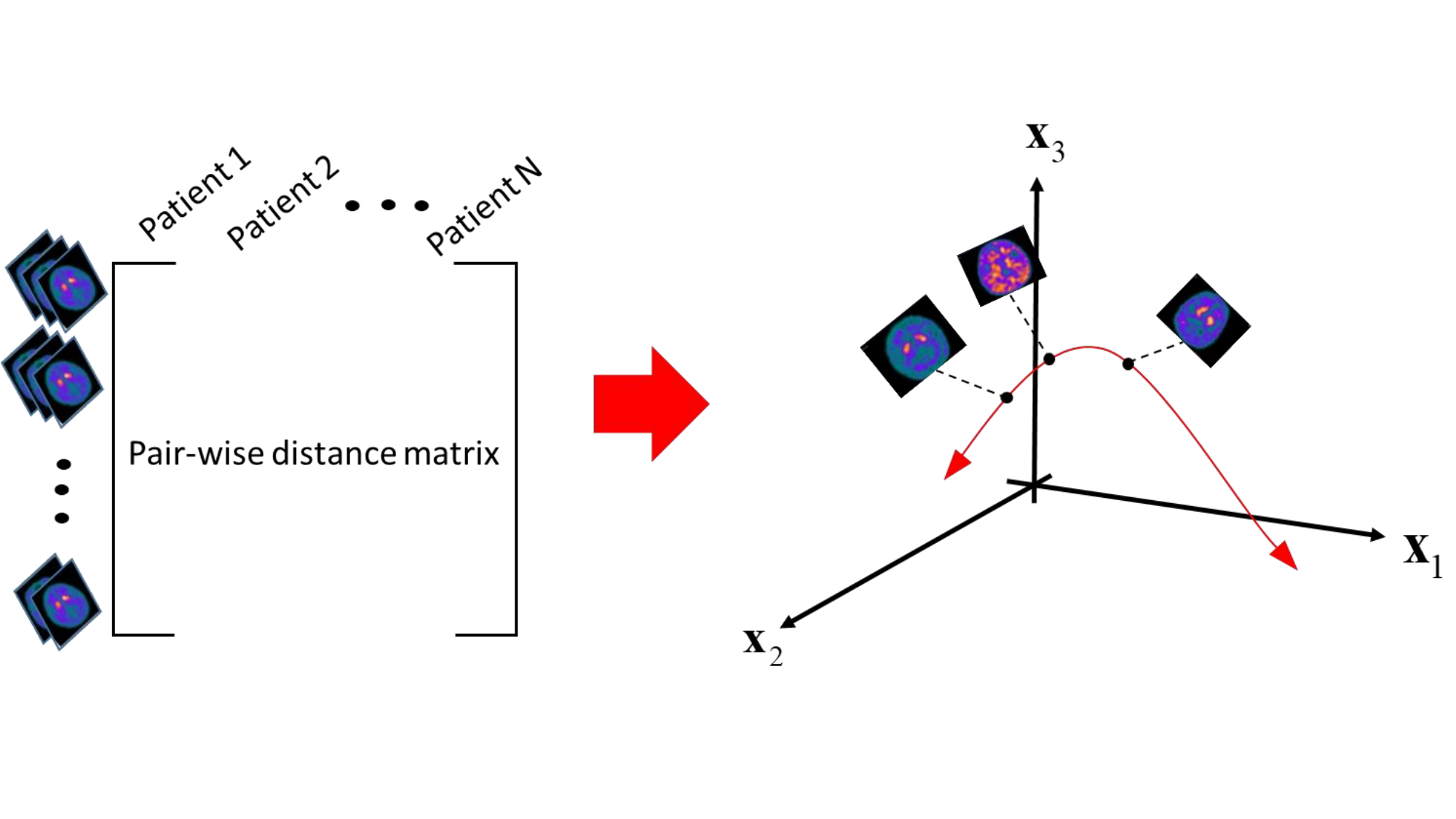}
    \captionsetup{font={normalsize}} 
    \caption{Compute pairwise of SPECT features distance and then mapping to low-dimensional space for image embedding.}
    \label{fig:image_embedding}
 \end{figure}

Our framework uses a graph model treating data points (samples) as nodes connected by edges with distances,  defined by a weighting scheme with 
$w_{ij}$ denoting distance of node $i$ to node $j$, for all $i,j=1,\cdots,n$. Given $n$ data point set $\left\{X_{i}\right \}_{i=1}^{n}$, where $X_{i} \in \mathbb{R}^{N \times N}$, $i = 1,2,...,n$. The data point is embedded to a manifold surfaces $\mathcal{M}$ in high-dimensional space and later is mapped into a lower dimensional space through a diffusion maps as shown in \textbf{Fig. \ref{fig:image_embedding}}. For illustration let an undirected graph $G(V,E,W)$, in which $V$ and $E$ are the set of vertices and edges, respectively, be defined that each node is connected by the weighted edges matrix $W=(w_{ij})$, where $w_{ij}$ is the weight of the edge $i$ and $j$ connecting.

 In diffuison maps, the weights between node $i$ and $j$ is usually defined through a kernel function  $\mathcal{K}(X_{i},X_{j})$  such as the Gaussian \cite{Porte08anintroduction} kernel with $w_{ij}=\mathcal{K}(X_{i},X_{j})$, where $\mathcal{K}$ is a positive and symmetric kernel matrix using Euclidean distance  $\left\| X_{i} - X_{j}\right\|^2 $ for the $X_{i}$ and $X_{j}$ metric

\begin{equation}
\mathcal{K}_{ij}=\mathcal{K}(X_{i},X_{j}) =\exp(-\frac{\| X_{i} - X_{j}\|^2}{\alpha}).
\end{equation}
    
\noindent The $\alpha$ is a scale parameter which measures the $X_{i}$ and $X_{j}$ in manifold neighborhood distance, 
we can construct a matrix with normalized rows entry with unit length as

\begin{equation}
P=D^{-1}\mathcal{K}.
\end{equation}

\noindent where $i$th element of degree matrix $D$ is computed by $d_{ii} = \sum_{j=1}^n \mathcal{K}_{ij} $ and denoted as $D =$
$\textbf{diag}{(d_{11},d_{22},...,d_{nn})}$. The resulting matrix $P$ is actually a  Markov transition matrix with every entry to be nonnegative and has all row sums to be equal to 1. The position of the data point $X_{i}$ and $X_{j}$ is connected on intrinsic manifold $\mathcal{M}$ by weights, so $p(X_{i},X_{j})$ is normalized by $\mathcal{K}_{ij}/d(x_{i})$  can be presented

\begin{equation}
P = (p(X_{i},X_{j})).
\end{equation}

\noindent If points $X_i$ and $X_j$ are similar, the distance between the two points is closer, and is a higher probability jumping from node $i$ to nearby node $j$. The similarity of the two data points can be  evaluated by the following distance measure

\begin{equation}
\begin{aligned}
D^{2}(X_{i},X_{j})&=\sum_{u \in X} \| p(X_{i},u) - p(X_{j},u) \|^2 \\
&=\sum_{k}\|p(X_{i},u) - p(X_{j},u) \|^{2} \\
&=\sum_{k}\|P_{ik} - P_{kj} \|^2.
\end{aligned}
\end{equation}

\noindent And let $Y_i$ be the $n \times 1$ distance vector with elements composed of distances from node $i$ to all nodes $j=1,\dots,n$ (including itself), mapped from the original data point $X_i, i=1,\dots,n$, namely

\begin{equation}
Y_{i}:= \begin{bmatrix} p(X_{i},X_{1})\\p(X_{i},X_{2})\\ \vdots
\\ p(X_{i},X_{n})
\end{bmatrix} = P_{i\cdot}^{T\quad}
\end{equation}

\noindent Note that the Euclidean distance between $Y_i$ and $Y_j$ can be expressed as

\begin{equation}
\begin{aligned}
\| Y_{i}-Y_{j} \|_{E}&=\sum_{k}\|p(X_{i},u) - p(X_{j},u) \|^{2} \nonumber    \\
&= \sum_{k}\|P_{ik} - P_{kj} \|^{2} 
\nonumber \\
&= D(X_{i},X_{j})^2.
\end{aligned}
\end{equation}

\subsection{Nystr$\ddot{o}$m's Out-Of-Sample Extension}
\label{susec:Nystr}
Manifold learning methods usually require recalculation of the the kernel matrix with new  added samples, and the above steps are repeated for the entire dataset. It becomes more difficult when new sample points are added sequentially. Nystr$\ddot{o}$m's out-of-sample extension \cite{Bengio04out-of-sampleextensions} allows the original sample to be extended by adding new samples and embedding them into the existing low-dimensional space to form the manifold geometry of the new sample points. The benefits  of Nystrom's out-of-sample procedure extending the new samples are illustrated as follows: (1) The compliance of machine learning is divided into training set and testing set standard verification steps; (2) When new samples are added, it is not necessary to repeatedly calculate diffusion distance of whole Markov matrix which also solves the problem of computation time; (3) Maintain the training set geometric structure from the original samples.

We can extend new sample point $X_{new} \in \mathbb{R}^{N \times N}$ from the validation set, and then recalculate the Euclidean distance and Kernel matrix,  the new element of Kernel matrix has an augmented vector $\mathcal{K}_{new}^T=(\mathcal{K}_{new,1},\dots,\mathcal{K}_{new,n})$ with 
\begin{equation}
     \mathcal{K}_{new,j}=\mathcal{K}(X_{new},X_{j}) =\exp(-\frac{\|  X_{new} - X_{j}\|^2}{\alpha}).
\end{equation}
The augmented  vector ${P}_{new}^{T}=D_{new}^{-1}\mathcal{K}_{new}^{T}$ is normalized vector of $\mathcal{K}_{new}^
T$ by dividing its row sum \\ $D_{new}=\sum_{j}^{n}\mathcal{K}_{new,j}$. So the augmented Markov transition matrix $P_{n+1}$ can be rewritten as
\begin{equation}
P_{n+1}=\begin{pmatrix}
P_{n} \\
P_{new}^{T}
\end{pmatrix},\mathcal{K}_{n+1}=\begin{pmatrix}
\mathcal{K}_{n} \\
\mathcal{K}_{new}^{T}
\end{pmatrix}
\end{equation}
Finally, project a new sample to the diffusion space of chosen $k$ eigenvector above by

\begin{equation}
    \psi_{l}(X_{new})=\frac{1}{\lambda_{l}}\sum_{j=1}^{n}p(X_{new},X_{j})\psi_{l}(X_{j}),l=1,...,k.
\end{equation}

\begin{figure}[H]
    \centering
    \includegraphics[width=0.7\linewidth]{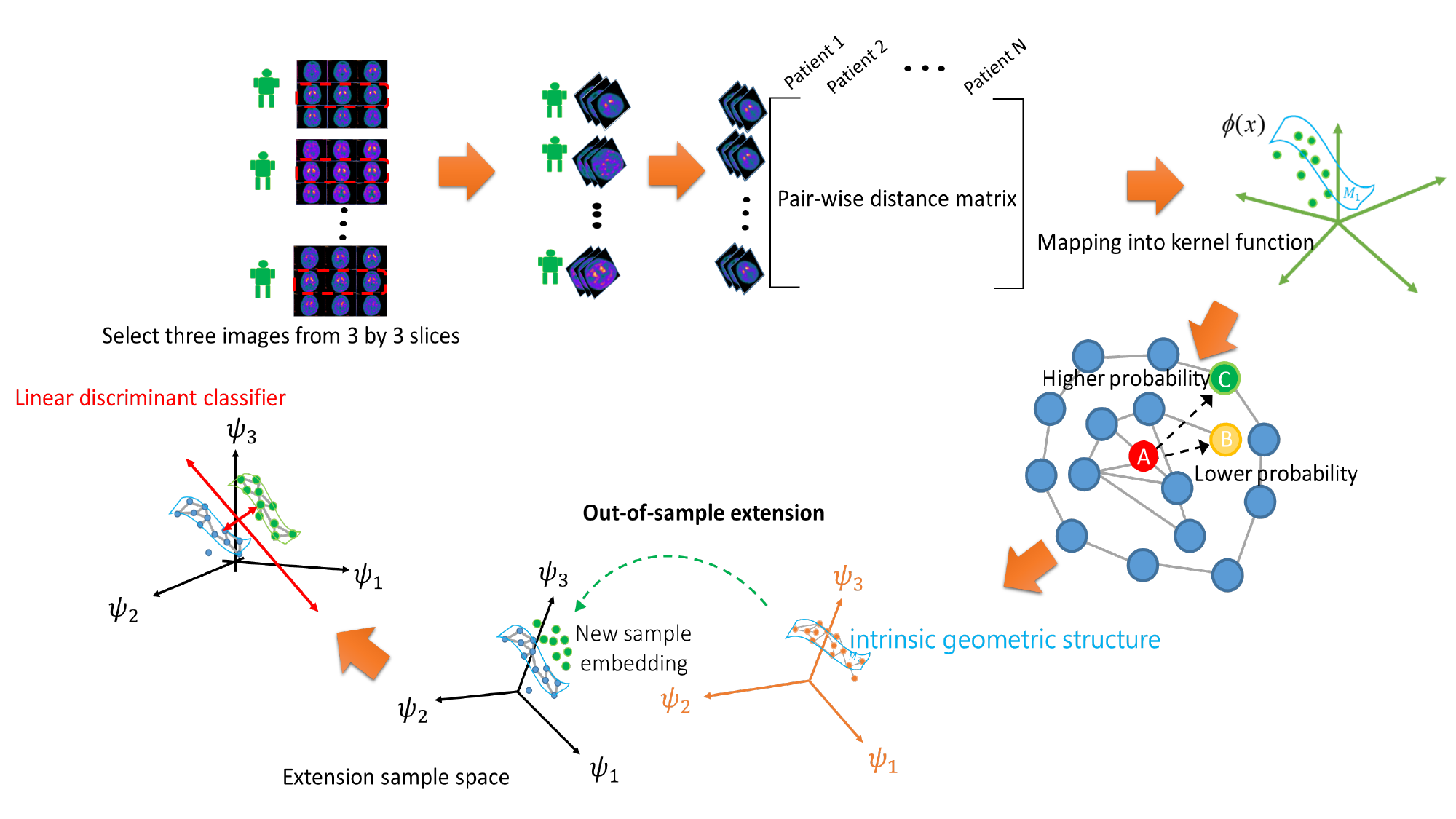}
    \captionsetup{font={normalsize}}
    \caption{The process of the high-dimensional SPECT data reduction uses out-of-sample porjects for classifications.}
    \label{fig:flow}
\end{figure}
Each point for each patient is projected onto the original trained low-dimensional diffusion space, and it goes on to use the selected classifier to carry on the classification as shown in \textbf{Fig. \ref{fig:flow}}. The algorithm for constructing the corresponding diffusion maps and out-of-sample extension is presented in \textbf{Algorithm. \ref{alg:Framwork}} below.

\begin{algorithm}[h] 
\caption{Diffusion Maps and out-of-sample extension algorithm.} 
\label{alg:Framwork} 
\begin{algorithmic}[1] 
\REQUIRE ~~\\ 
Data of images $\{X_{i}\}_{i=1}^{n} \in \mathbb{R}^{N \times N}$;\\

\ENSURE ~~\\ 
Projected new coordinate vector $Y'_{new} \in \mathbb{R}^{N \times k}$;
\STATE \textbf{Normalized data:} $\tilde{X}_{i},i=1,2,..,n.$ \\
 $\tilde{X}_{i}=(X_{i}-\bar{X})/S(X_{i})$, which $\bar{X}_{i}= \frac{1}{N}\sum_{j,k}X_{i}(j,k)$\\
 $S({X}_{i})=\frac{1}{N^{2}-1}\sum_{j,k}(X_{i}(j,k)-\bar{X}_{i}(j,k))^{2}$; 
\label{ code:fram:extract }
\STATE \textbf{Construct Kernel matrix:}$K=(K_{i,j}), K_{i,j}=K(\tilde{X}_{i},\tilde{X}_{j})$; 

\STATE\textbf{Build Markov matrix:}$P=D^{-1}K$,\\
$D = \textbf{diag}{(d_{11},d_{22},...,d_{nn})}$, where $d_{ii} = \sum_{j=1}^{n}K_{ij}$;
\label{code:fram:trainbase}
\STATE \textbf{Spectral decomposition $P$ matrix with corresponding eigenpairs
  $\{\lambda_{j},\psi_{j} \}_{j=1}^{n}$ as coordinate:}\\
 
$Y_{i}^{\prime}$ $^T$ $:=(\lambda_{1}\psi_{1}(i),\lambda_{2}\psi_{2}(i)...,\lambda_{k}\psi_{k}(i))$
\label{code:fram:add}
\STATE \textbf{Out-of-sample extension}
\label{code:fram:classify}
\STATE \textbf{Compute new extension vectors:}\\
$P_{new}^{T}=d_{new}^{-1}K_{new}^{T}$, where $K_{new}^{T}$=
$(K_{new,1},...,K_{new,n})$; 
\STATE  \textbf{Repeat Step 1 to Step 4 and compute extented Markov matrix:}\\
$P_{n+1}^{T}=(P_{n},P_{new}^{T})$, where $K_{n+1}^{T}=(K_{n},K_{new}^{T})$;
\label{code:fram:select}
\STATE \textbf{Project new samples on eigenvecotrs as coordinate:}\\
 $\psi_{l}(X_{new})=\frac{1}{\lambda_{l}}\sum_{j=1}^{k}p(X_{new},X_{j})\psi_{l}(X_{j}),l=1,...,k$;\\

\RETURN  $\psi_{l}(X_{new}),l=1,...,k$; 
\end{algorithmic}
\end{algorithm}

\begin{figure}[H]
    \centering
    \includegraphics[width=0.95\linewidth]{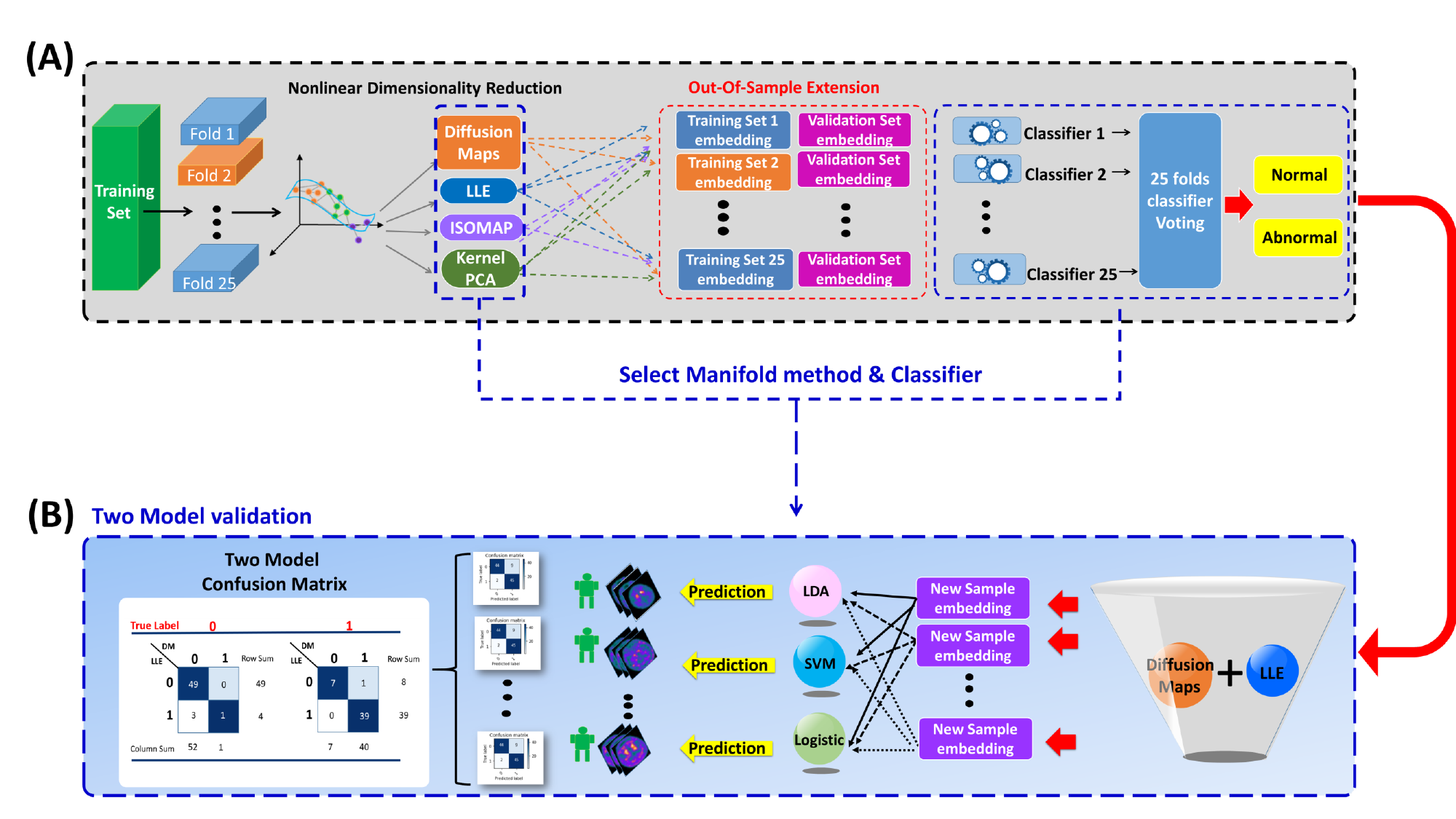}
    \captionsetup{font={normalsize}}
    \caption{The workflow of the two steps model diagnosis architecture with twenty-five folds cross-validation. 
    (A) Procedure for nonlinear dimension reduction through Manifold method and diagnosis with given classifier. (B) Procedure for new test samples embedding and diagnoses with DM and LLE as well as the corresponding two model confusion matrix.}
    \label{fig:work_flow}
\end{figure}

\section{Experiments}

In this work, we implement the same data preprocessing procedure in our clinical data KCGMH-TW and PPMI dataset. Our approach mainly consists of two steps as demonstrated in  \textbf{Fig. \ref{fig:work_flow}}:

\begin{itemize}
  \item  
    First step: several manifold learning methods are used for dimension reduction on the training sets for low-dimensional embedding, and then out-of-sample extension is applied to the new samples of the validation sets\cite{Raeper2018}. Next, classify the embedded samples and aggregate the ensemble classifiers with the twenty-five voting results to making prediction on the  corresponding test set. 
  \item 
   Second step: according to the results of the first step, we use the DM and the classifier with the best performance to predict new test samples. Each patient is diagonalized through the twenty-five classification voting procedure and the prediction results  with different manifold learning methods can be compared through the two model confusion matrices.   
  
\end{itemize}

In the first step, three classifiers are compared, namely LDA, SVM and Logistic regression. The classifier with the best performance is chosen to enter the second step process. In this work, the DM parameter is set to be $\alpha=8$ and time step $t=1$ in Markov transition matrix $P^{(t)}$. In out-of-sample step, there is no need to recompute the entire diffusion matrix, simply project the new samples to the original chosen 
DM space which will be used as the feature space for diagonsis analysis.

\subsection{Two Steps Model Ensemble and Classifer Selection}
\label{ensemble classifer select}

In order to  compare diagnosis results using DM with those by manifold learning methods such as LLE, ISOMP, KPCA, and classifiers such as LDA, SVM, Logistic Regression in another scenario, we compare the performances of DM, LLE, ISOMAP, and KPCA  diagonalized results based on the twenty-five folds classifications under four low-dimensionality (30, 100, 200 and 300) with different classifiers. Overall accuracy averages based on the twenty-five folds predictions after voting with the three classifiers are presented in \textbf{Table. \ref{table_classifier} }.  It can be seen that (DM, LDA) pair with 200 dimensionality seems to be the best. An interesting observation on the accuracy of the four methods, is that after dimension reduction, the linear discriminant classifier (LDA) works better than the other two non-linear classifiers. 
The performances of the DM and LLE are superior to the other two methods, therefore in the following we use DM and LLE with LDA for comparisons of the performances for testing the new 100 samples. 

\begin{table}[h]

 \centering
  \fontsize{10}{8}\selectfont
  \captionsetup{font={normalsize}}
  \caption{Performance comparisons of the four manifold learning methodology and three ensemble classifiers on KCGMH-TW }
  \label{tab:performance_comparison}
    \begin{tabular}{|c|c|c|c|c|c|c|c|c|c|c|}
    \hline
    \multirow{2}{*}{Method}&
    \multicolumn{4}{c|}{LDA}&\multicolumn{3}{c|}{SVM}&\multicolumn{3}{c|}{Logistic}\cr\cline{2-11}
    & Dimension &\textbf{Acc.}&\textbf{Sens.}&\textbf{Spec.}&\textbf{Acc.}&\textbf{Sens.}&\textbf{Spec.}&\textbf{Acc.}&\textbf{Sens.}&\textbf{Spec.}\cr
    
    \hline
    \hline
    \multirow{4}{*}{DM} &30&0.88&0.93&0.82&0.86&0.93&0.79&0.85&0.90&0.80 \\
	&100&0.89&0.94&0.84&0.86&0.89&0.83&0.88&0.90&0.86\\
	&200&{\bf 0.90}&{\bf0.91}&{\bf 0.91}&0.85&0.86&0.84&0.87&0.88&0.86\\
	 &300&{\bf 0.91}&{\bf 0.91}&{\bf 0.91}&0.87&0.92&0.82&0.88&0.92&0.84\\ \hline
	 \multirow{4}{*}{LLE} &30&0.84&0.87&0.81&0.83&0.88&0.77&0.81&0.84&0.79\\
	 &100&0.89&0.96&0.81&0.83&0.88&0.77&0.81&0.84&0.79\\
	 &200&0.89&0.93&0.83&0.87&0.93&0.81&0.85&0.87&0.82\\
	 &300&0.89&0.93&0.83&0.87&0.81&0.93&0.85&0.87&0.82\\ \hline
	 \multirow{4}{*}{ISOMAP} &30&0.80&0.81&0.79&0.81&0.83&0.79&0.79&0.77&0.81\\
	 &100&0.86&0.89&0.84&0.85&0.86&0.84&0.85&0.84&0.85\\
	 &200&0.87&0.89&0.85&0.86&0.86&0.85&0.87&0.87&0.87\\
	 &300&0.87&0.89&0.85&0.86&0.86&0.85&0.87&0.87&0.87\\ \hline
	 \multirow{4}{*}{KPCA} &30&0.79&0.80&0.79&0.76&0.80&0.71&0.75&0.85&0.70\\
	 &100&0.83&0.83&0.82&0.77&0.82&0.71&0.77&0.83&0.70\\
	 &200&0.86&0.90&0.81&0.79&0.85&0.72&0.78&0.84&0.71\\
	 &300&0.87&0.93&0.81&0.80&0.89&0.73&0.78&0.84&0.71\\ \hline
    
    \end{tabular}
    \label{table_classifier}
\end{table}

\vspace{10pt} For each test patient, based on the two-step procedure above, there are twenty-five ensemble votes for prediction, then we can calculate the proportion of machine predicted abnormal probability: $p_{k} =\frac{V_{k}}{\sum_{k=1}V_{k}}$, with $V_k$ denoting the $k$th  voting results, where $V_k= 0$ or $1$ representing normal or abnormal respectively, $k=1,\ldots,25$. So we build a threshold to (0,1) \emph{e.g.,}($\mathbbm{1}_{\{p_{k}> 0.5\}}$) for testing the voting prediction. 

\subsection{Classification}

We compare the performances of the two-step manifold approach with 2D-CNN, 3D-CNN on KCGMH-TW  and PPMI datasets in twenty-five folds predict. And then we choose the best performing 200 dimensions as a baseline and compare diﬀerent well-known CNN models such as AlexNet\cite{Krizhevsky12imagenetclassification}, VGG-16\cite{Simonyan2015}, VGG-19 and Deep Convolution Network (DCNN). We implement the KCGMH-TWdataset with three SPECT images from $9\times 9$ slices as a high-order brain image tensor as input to the 3D-Convolutional Neural Network model\cite{3D_CNN}. The volumes of SPECT have a size of $3 \times128\times 128\times 3$ voxels. For the CNN approach, we set 193 images as the test set and 100 images in the manifold method test in \textbf{Table.~\ref{Group table}}.

Evaluate out-of-sample extension to new samples, we use Frobenius norm distance to measure testing sample point embedding on low-dimensional space. In KCGMH-TW (three images) case, three images show that between the training set over to embed testing set significant effect classification accuracy like a \textbf{Table.~\ref{Group table}}.

We consider the early abnormal in our KCGMH-TW for training and testing (testing = 100) and implement our procedure in the PPMI dataset (testing = 197). The classification results are reported in Table.~\ref{Group table}. DM+LDA method generally achieves the best performance on two datasets with single and three types. Compare to our approach methods, 2D-CNN often have excellent prediction than 3D-DCNN such that VGG-16 in PPMI is 93\% and DCNN in KCGMH-TW single image is 87\% but also has higher variability of 0.13 and 0.22. The comparison of our method DM+LDA in PPMI single image and KCGMH-TW three images classiﬁcation has average accuracy 98\% and 90\% performance with lower variability is 0.02 and 0.05.

\begin{table}[H]
\centering
\captionsetup{font={normalsize}}
\caption{Binary classification in different datasets}
\label{tab2}

\setlength{\tabcolsep}{1mm}{
\begin{tabular}{lllllllll}

\toprule[2pt]
\textbf{Datasets}&\textbf{Single image} & \textbf{Model} & \textbf{Acc.} &\textbf{ Prec.} &\textbf{ Senc.} & \textbf{Spec.} \\

\hline
\hline
 PPMI&HC vs PD & DCNN & 0.89 ($\pm0.54$) & 0.92 & 0.85 & 0.90\\ 
 && AlexNet & 0.88 ($\pm0.27$)& 0.89 & 0.78 &0.92\\
 && VGG-16 & 0.93 ($\pm0.13$)& 0.94 & 0.95 &0.92\\
 && VGG-19 & 0.89 ($\pm0.21$)& 0.91 & 0.93&0.86\\
 && Diffusion Maps + LDA & \textbf{0.98} ($\pm0.02$) & 0.98 & 0.96 & 0.97 \\

KCGMH-TW&Normal vs Abnormal& DCNN & 0.87 ($\pm0.22$) & 0.88 & 0.89 & 0.84\\ 
 && AlexNet & 0.83 ($\pm0.35$)& 0.86 & 0.84 &0.80\\
 && VGG 16 & 0.85 ($\pm0.44$)& 0.84 & 0.89 &0.76\\
 && VGG 19 & 0.87 ($\pm0.11$)& 0.88 & 0.92&0.81\\
 && Diffusion Maps + LDA & 0.86 ($\pm0.04$) & 0.85 & 0.93 & 0.77 \\
\hline

\toprule[2pt]
\textbf{Datasets}&\textbf{Three images} & \textbf{Model} & \textbf{Acc.} &\textbf{ Prec.} &\textbf{ Senc.} & \textbf{Spec.} \\
\hline
\hline
KCGMH-TW&Normal vs Abnormal & 3D-DCNN & 0.82 ($\pm0.10$) & 0.83 & 0.84 & 0.79 & \\
&& Diffusion Maps + LDA & \textbf{0.90} ($\pm0.05$) & 0.88 & 0.95 & 0.84\\
\hline

\toprule[2pt]
\label{Group table}
\end{tabular}}
\end{table}

\begin{figure}[ht]
\centering
\subfigure[PPMI (single image)]{\includegraphics[height=5.2cm,width=6.5cm]{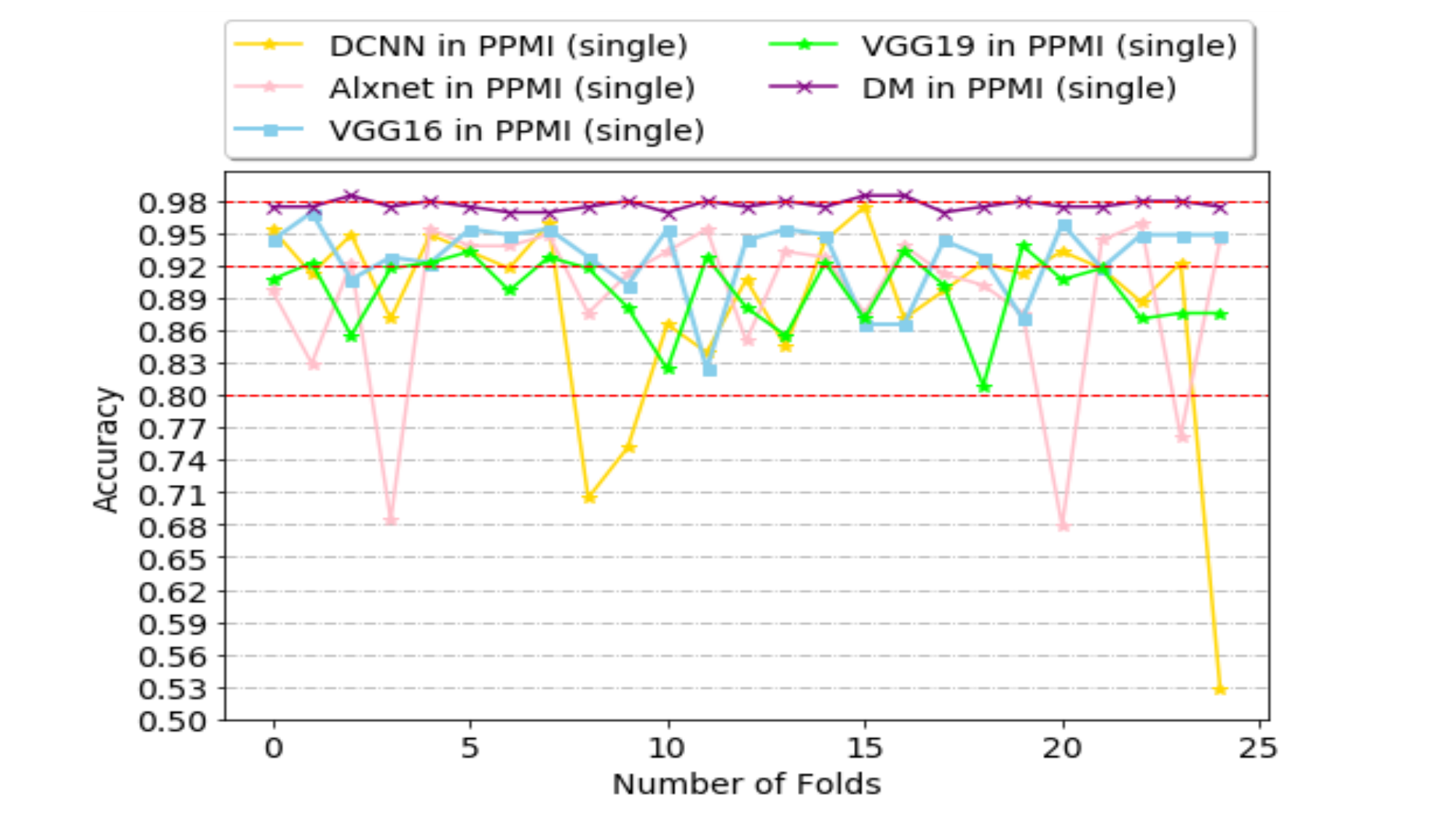}}
\hspace{4mm}
\subfigure[KCGMH-TW (single image)]{\includegraphics[height=5.6cm,width=7.5cm]{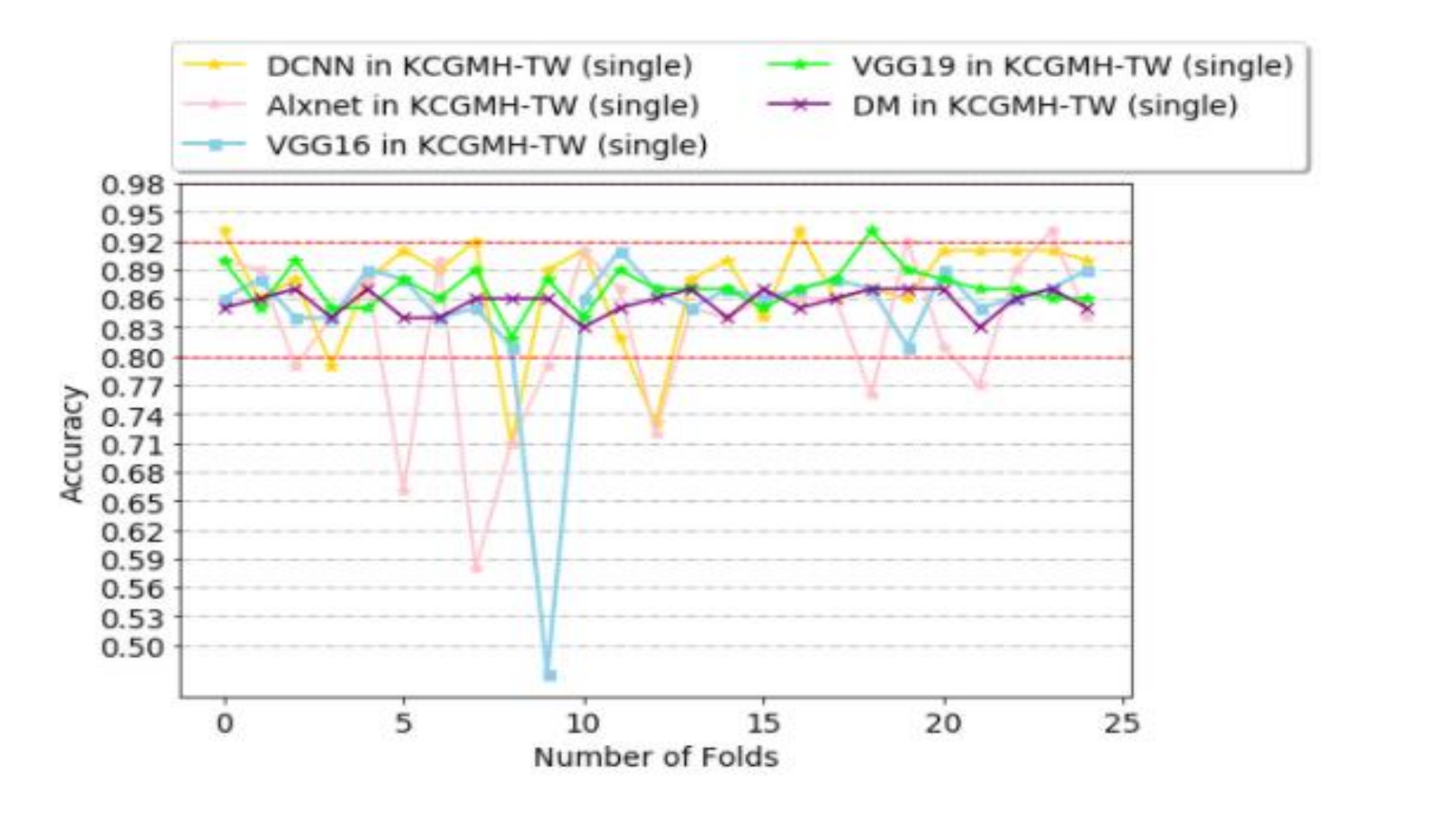}}
\captionsetup{font={normalsize}}
\caption{The performace of DM and CNN comparison on PPMI and KCGMH-TW dataset.}
\label{fig:Overall_model_plot}
\end{figure}

From the overall results in \textbf{Fig. \ref{fig:Overall_model_plot}}, we draw DM+LDA and CNN methods in twenty-five folds situation. In \textbf{Fig. \ref{fig:Overall_model_plot}(a)(b)}, display that we only use a single PPMI image to predict 193 patients, DM+LDA has robust performance than VGG-16, VGG-19, AlexNet, and DCNN. The PPMI dataset has high image quality compared to our clinical KCGMH-TW dataset. To reduce effects of the noise in our KCGMH-TW, we also use three images for DM+LDA for comparisons. We find that DM+LDA procedure has more robust results with three images, see \textbf{Fig. \ref{fig:Overall_model_plot}(c)} and \textbf{Fig. \ref{fig:Overall_model_plot}(d)}. It may be due to that three images have included the most obvious symmetrical strata in high order tensor space.

\begin{figure}[ht]
\centering
\includegraphics[width=3.4in]{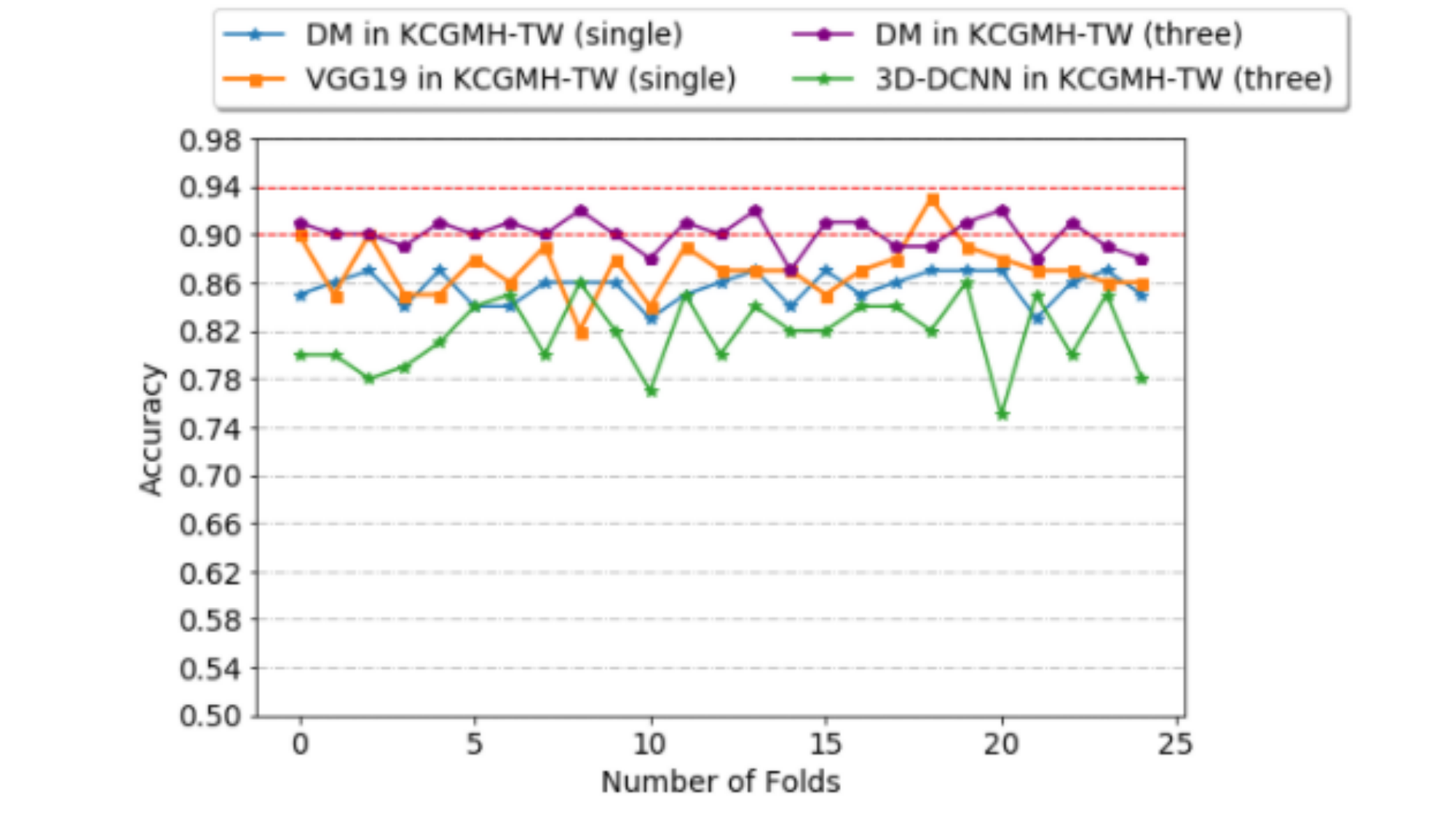}
\captionsetup{font={normalsize}}
\caption{Performance of DM in single image, three images and 3D-DCNN methods}
\label{fig:DM_single_triple_3DCNN}
\end{figure}

\section{Diagnosis and Discussion}

In this section, we focus on machine prediction on the 100 test patients and diagnosis of misclassification samples. Our diagnostic framework as exhibited in  \textbf{Fig. \ref{fig:work_flow}} (B), provides a general interpretable misclassification table for diagnosis in the final step with the confusion table. It is helpful to see whether an incorrect diagnosis is dut to the patient's age or not.

\subsection{Two Model Confusion Matrices}
\label{Two model confusion matrices}

\begin{figure}[H]
\centering
\includegraphics[width=5.5in]{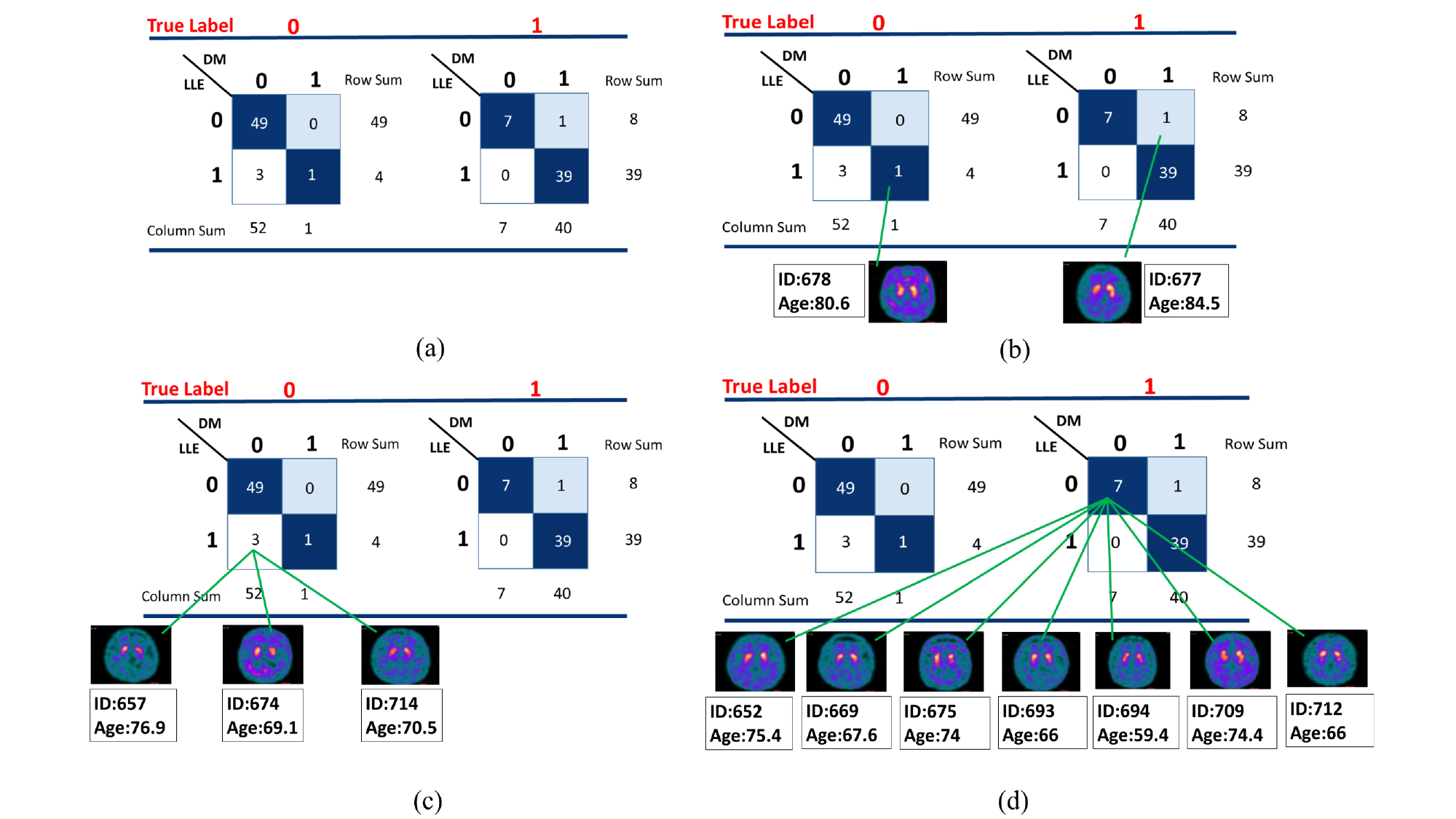}
\captionsetup{font={normalsize}}
\caption{Two-model confusion matrices}
\label{fig:two_model_confusion_OK}
\end{figure}

From Section~\ref{ensemble classifer select} , it is obvious that DM and LDA classifier have high performance in classification after the twenty-five folds voting. As LLE also has quite good performance with  the twenty-five folds procedure, we examine the diagnosis results based on the two models and form a two-model confusion matrix. First, we set the doctor label as true label (i.e., two classes) and the ensemble DM and LLE diagnoses in two classes as in  \textbf{Fig. \ref{fig:two_model_confusion_OK}(a)} and then check how many misclassifications under the true label normal or abnrmal.

The diagonal elements of the two model confusion matrix table are the numbers of correct predictions (positive  is with label \textbf{1}, negative is with \textbf{0}) for the DM and LLE ensembles respectively. The numbers of misclassified samples are shown in the non-diagonal sub-table. 
For example in the left table in \textbf{Fig. \ref{fig:two_model_confusion_OK}(b) } left column is ture label of normal, DM and LLE predict ID:678 patient is early abnormal. In the right column \textbf{Fig. \ref{fig:two_model_confusion_OK}(b) }, ID:677 is predicted  by LLE and abnormal is predicted incorrectly as normal by LLE and correctly as abnormal by DM, while ID:678 is predicted incorrectly as normal by both models. In the left column of the table \textbf{Fig. \ref{fig:two_model_confusion_OK}(c) }  as  abnormal, three of the patients with true label as normal are diagnosed correctly as normal (ID: 657, 674, 714 ) by DM and incorrectly as abnormal.   

\begin{figure}[H]
\centering
\includegraphics[width=4in]{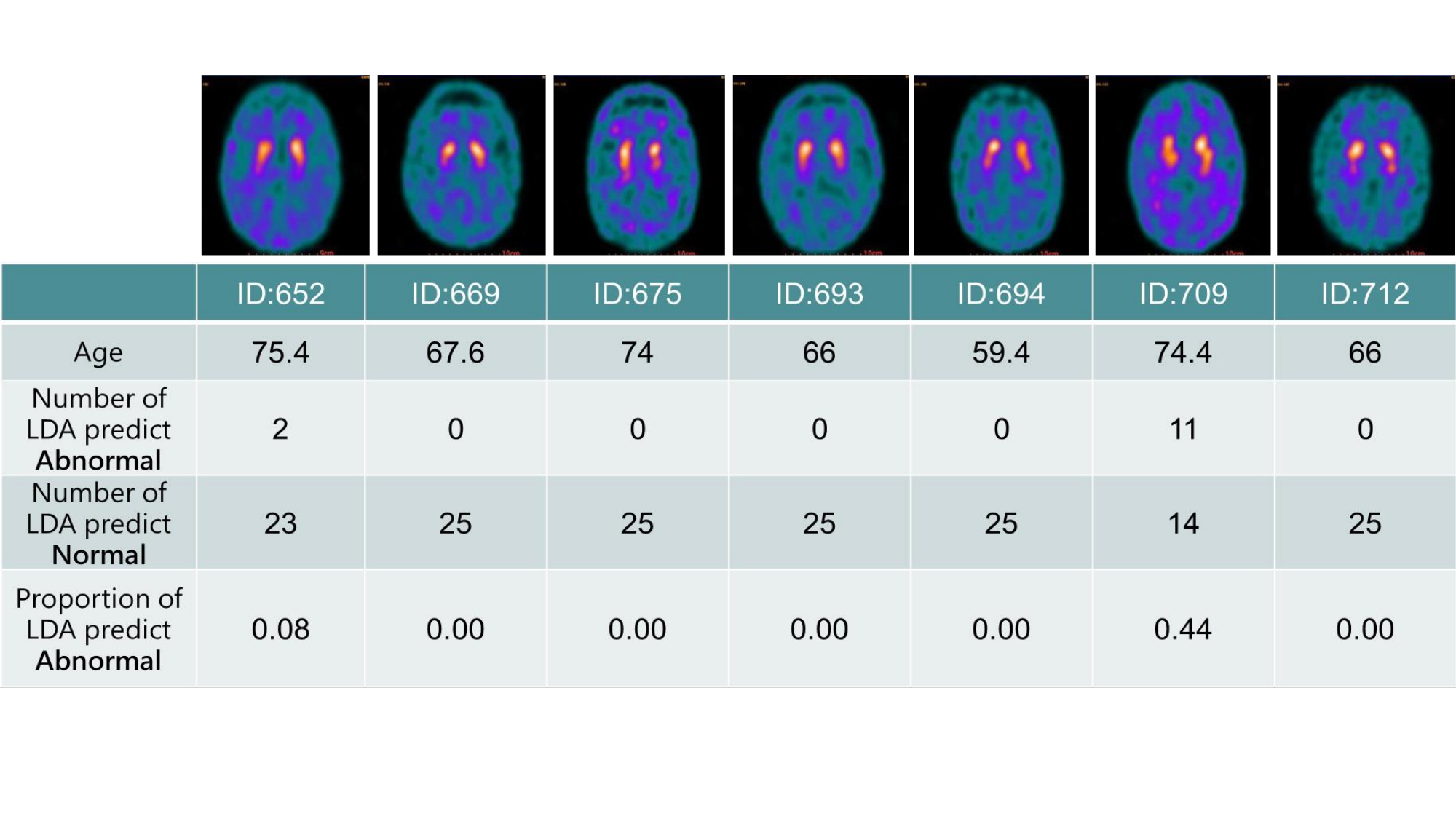}
\captionsetup{font={normalsize}}
\caption{The diagnosis table for the seven false negative samples on their age and proportion of the twenty-fold voting results.}
\label{Table:table_11}
\end{figure}

\noindent\textbf{Diagnosis Table:} \hspace{8pt}Sometimes we care more about false negative situation as the seven cases as indicated in  \textbf{Fig. \ref{fig:two_model_confusion_OK}(d)}. For these misclassifications, we can provide more detailed information for the doctor's diagnosis table and the probability of LDA matching as demonstrated in  \textbf{Table. \ref{Table:table_11}}, in order to identify early abnormal in the true designation of three classes. We can also check if our model predicts an error due to high age or shrinking striatum abnormal. For example, in the case of the misclassification ID:709, whose age is 74.4 higher than the average age of our dataset of 68.3.

\subsection{Visualization}

\begin{figure}[ht]
    \centering
    \includegraphics[width=0.8\linewidth]{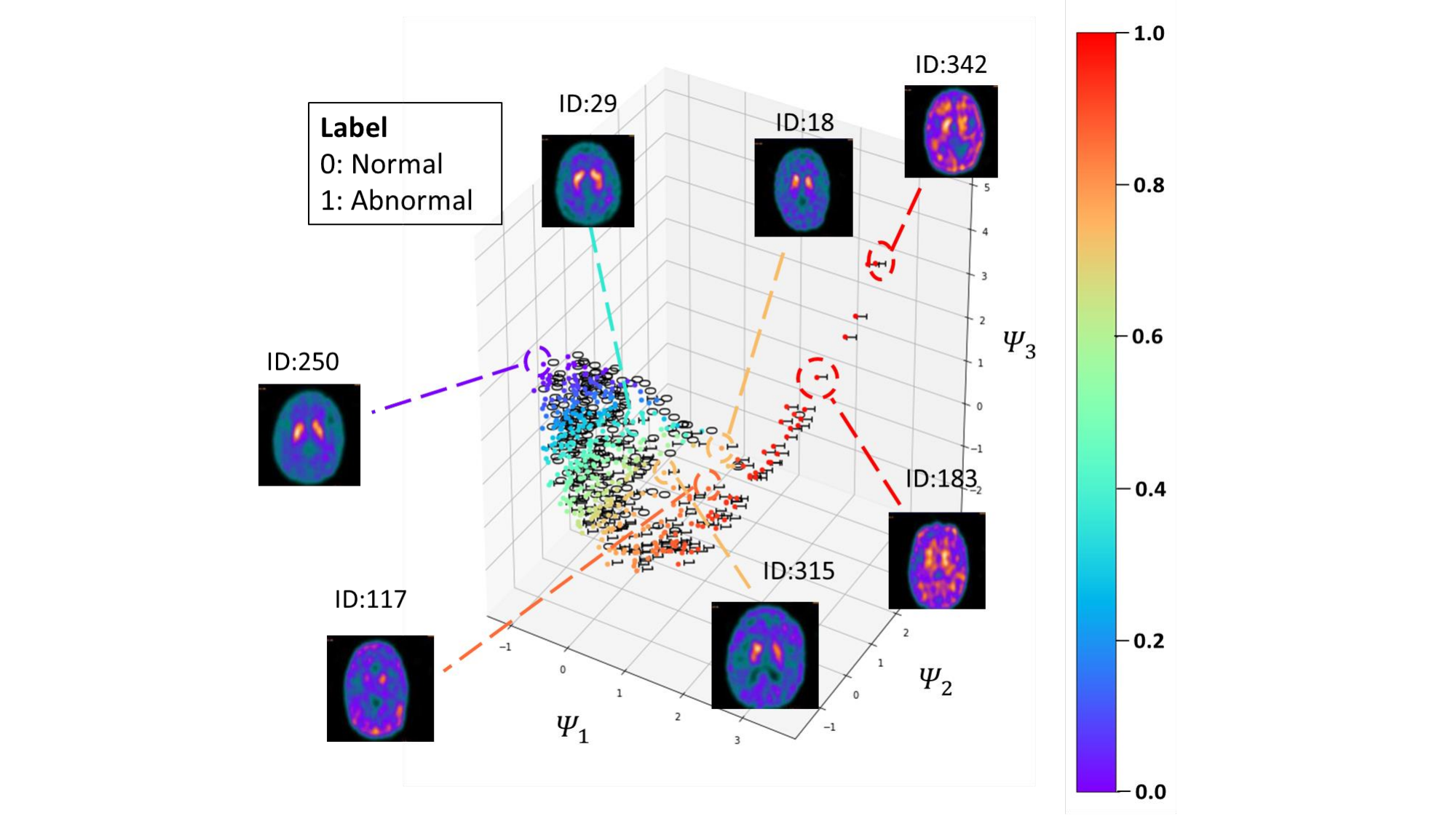}
    \captionsetup{font={normalsize}}
    \caption{Trajectory of 630 SPECT images embed in three-dimentional space and annotated the true label on each data points. The manifold is colored by
    SPECT similarity distance of DM and eigenvectors mapping.}
    \label{fig:time_step2}
\end{figure}

Traditional analysis of the SPECT image is for the physician to visualize the symmetry of the left and right striatum. However, the early abnormal symptoms is more controversial.
To find mild changes of Parkinson's disease, we use DM for dimension reduction on 630 original SPECT images in low-dimensional space, and visualize the embedding test samples in the corresponding three-dimensional manifold using the first three eigenvector $\psi_1,\psi_2, \psi_3$ with the largest three eigenvalues.  They are shown to be like an U curve with abnormal samples appeared on the right side of the curve as shown in \textbf{Fig. \ref{fig:time_step2}}.  The color from dark purple to dark red indicates the distance between sample points of each patient. Moreover, we label the true diagnosis of each patient (two classes) to examine the spatial status. The data point (ID:342) with the largest distance from the origin on the right extreme is the most serious abnormal case. In contrast, the point (ID:250) on the left extreme is normal, and the early abnormal appears in the middle position, such as ID: 18.

\section{Conclusion}

In addition to maintaining excellent geometric structure in low dimensions, DM is computationally less expensive than deep learning methods and does not require too many parameter adjustments.

In section 4.2, the average accuracy of DM +LDA classification for KCGMH-TW normal vs. abnormal (three images) is up to
90\%  and up to 98\% for PPMI dataset (single). The overall
performance on KCGMH-TW is quite accurate and robust with lower variation than other deep learning methods. For understanding precisions of diagnoses of the two manifold learning methods, in section 5.1 we have constructed confusion matrix of the two most compatible models.  We examined those samples which were misclassified by our methods in more details and found out that many were diagnosed as early abnormal by the doctors, and sometimes there were differences on the diagnoses among doctors. 
Through the confusion matrix we are able to provide some explainable reasons for the diagnosis  discrepancies  among the doctors and  our methods. Finally, through DM, we can embed the sample images into lower dimensional space and visualize how the normal and abnormal sample images scatter around in the three-dimensional eigenspace corresponding to  the largest three eigenvalues of the DM  method. In future works, we will include other existing diagnosis variables with extracted features to cross-examine the diagnosis results and see if we may improve the diagnosis accuracy further. Moreover, we will investigate, if more than 3 slices of images for each subject are used,  it would be helpful for detection of early abnormal cases with more information.

%
%
%
%

\end{document}